\documentclass{article}

    \PassOptionsToPackage{numbers, compress}{natbib}

    \usepackage[final]{mlsafety_neurips_2022}

\usepackage[utf8]{inputenc} 
\usepackage[T1]{fontenc}    
\usepackage{hyperref}       
\usepackage{url}            
\usepackage{booktabs}       
\usepackage{amsfonts}       
\usepackage{nicefrac}       
\usepackage{microtype}      
\usepackage{xcolor}         
\usepackage{graphicx}
\usepackage{float}
\usepackage{amssymb}
\usepackage{amsmath}

\title{The Expertise Problem: \\Learning from Specialized Feedback}

\author{%
  Oliver Daniels-Koch\\
  Center for Human-Compatible AI \\
  UC Berkeley \\
  \texttt{danielskoch.oliver@gmail.com} \\
   \And
  Rachel Freedman\\
  Center for Human-Compatible AI \\
  UC Berkeley \\
  \texttt{rachel.freedman@berkeley.edu} \\
}

\begin{document}

\maketitle
\begin{abstract}
    Reinforcement learning from human feedback (RLHF) is a powerful technique for training agents to perform difficult-to-specify tasks. However, human feedback can be noisy, particularly when human teachers lack relevant knowledge or experience. Levels of expertise vary across teachers, and a given teacher may have differing levels of expertise for different components of a task. RLHF algorithms that learn from multiple teachers therefore face an \textit{expertise problem}: the reliability of a given piece of feedback depends both on the teacher that it comes from and how specialized that teacher is on relevant components of the task. Existing state-of-the-art RLHF algorithms assume that all evaluations come from the same distribution, obscuring this inter- and intra-human variance, and preventing them from accounting for or taking advantage of variations in expertise. We formalize this problem, implement it as an extension of an existing RLHF benchmark, evaluate the performance of a state-of-the-art RLHF algorithm, and explore techniques to improve query and teacher selection. Our key contribution is to demonstrate and characterize the expertise problem, and to provide an open-source implementation for testing future solutions. \footnote{Code for all experiments is available \href{https://github.com/oliveradk/BPref/tree/multi-teacher}{here}.}
\end{abstract}

\section{Introduction}
AI systems are typically trained to maximize a pre-specified objective but, for many tasks, defining an objective that fully captures the intentions of the designer is prohibitively difficult. One promising alternative to manually specifying objectives is reinforcement learning from human feedback (RLHF). In RLHF, human teachers compare pairs of trajectories against each other, the comparisons are used to train a reward model, and the reward model provides reward signal to further train the reinforcement learning agent \citep{christiano_deep_2017,lee_pebble_2021,lee_b-pref_2021}. RLHF has proven effective for training models on tasks that require human judgement, including summarizing texts and promoting helpfulness, honesty, and harmlessness in large language models \citep{stiennon_learning_2020, wu_recursively_2021, askell_general_2021, ouyang_training_2022}.

RLHF methods assume  all human feedback comes from a single human teacher. However, these methods typically require querying many teachers to gather sufficient training data, and all teachers are not equal - they vary in their ability to evaluate system behavior. For example, \citet{ziegler_fine-tuning_2020} finds that crowd sourced feedback often disagrees with the paper authors' feedback. An individual teacher's ability can also vary across queries. For example, \citet{stiennon_learning_2020} trains human teachers at different summarization domains, producing evaluators that ``specialize'' in evaluating either Reddit or CNN summaries. In the future, we expect RLHF to be applied to increasingly complex and compound tasks, where evaluators vary in their expertise across task components. RLHF methods may need to actively select teachers based on their expertise to ensure reliable feedback on different tasks. We call the problem of learning from multiple, specialized teachers the \textit{expertise problem}.  

Our contributions are as follows. We propose an abstract problem formulation of the expertise problem. Under this formulation, we test PEBBLE \citep{lee_pebble_2021}, a state of the art RLHF algorithm. We find that naive extensions of PEBBLE perform poorly, but that simple modifications, including selecting the most expert teacher and making intra-domain queries, can significantly improve performance. Our key contribution is to formalize and characterize the expertise problem, and to demonstrate the insufficiency of current methods. We hope that this formalism, demonstration, and open-source implementation will pave the way for more advanced RLHF methods that are robust to teacher variance and specialization.

\section{Preliminaries and Related Work}
\subsection{Reinforcement Learning from Human Feedback}
Reinforcement learning from human feedback (RLHF) uses comparison feedback to learn a model of a reward function, and uses the learned reward function to train reinforcement learners. As in traditional RL, at each timestep $t$ the learner observes a state $s_t$, takes an action $a_t$, and transitions to a new state $s_{t+1}$. These interactions produce $k$-length trajectories $\sigma \in \Sigma$, where $\sigma = \left( (s_1, a_1), ..., (s_k, a_k)\right) \in (\mathcal{S} \times \mathcal{A})^k$.

In RLHF, these trajectories are presented to human teachers as binary comparisons $(\sigma_1, \sigma_2)$, and teachers select which trajectory they prefer~\cite{christiano_deep_2017,lee_pebble_2021}. These teacher preference labels are represented by a distribution $\mu$ over $\{1, 2\}$ and stored in a database $\mathcal{D}$. RLHF methods typically model humans as Boltzmann-rational decision-makers, where the noisiness of their decisions is modulated by a ``rationality'' parameter $\beta$~\cite{ziebart2010modeling, jeon_reward-rational_2020}. The probability that a Boltzmann-rational decision-maker prefers trajectory $\sigma_1$ to $\sigma_2$ is:

\begin{equation}
    P[\sigma_1 \succ \sigma_2; \beta] = \frac{\exp(\beta \cdot r(\sigma_1))}{\exp(\beta \cdot r(\sigma_1)) + \exp(\beta \cdot r(\sigma_2))}
\end{equation}

where $r$  is the ``true'' reward function ($r(\sigma) = \sum_{n=1}^{k} r(s_n, a_n)$). A reward model $\hat{r}$ is trained to approximate the underlying reward function by minimizing the cross entropy between the teacher labels and the distribution $\hat{P}$ (given by substituting $\hat{r}$ for $r$):

\begin{equation}
    loss(\hat{r}) = \sum_{(\sigma_1, \sigma_2, \mu) \in \mathcal{D}}{\mu(1) \log{\hat{P}[\sigma_1 \succ \sigma_2]} + \mu(2)\log{\hat{P}[\sigma_2 \succ \sigma_1]}}
\end{equation}

The agent is simultaneously trained using the approximated reward function $\hat{r}$. 

\citet{christiano_deep_2017} uses A2C \citep{mnih_asynchronous_2016} and TRPO \citep{schulman_trust_2017}  to train the agents and preferentially samples queries that the reward model has greater uncertainty over. \citet{lee_pebble_2021} proposes PEBBLE, which uses SAC \citep{haarnoja_soft_2018} in conjunction with a self-supervised exploration bonus and trajectory relabeling to improve upon these results. Since PEBBLE is the current state-of-the-art, we'll use it in our experiments.

\citet{lee_b-pref_2021} benchmarks RLHF algorithms, using algorithmic teacher models to generate synthetic feedback data at scale. Crucially, they find that the optimal query sampling strategy varies across teacher feedback models, suggesting that identifying and leveraging differences between teachers may improve learner performance. Defining the teacher models algorithmically allows them to isolate these effects of differences in teacher decision-making, so we will use this strategy as well. 

\subsection{Estimating Expertise}
Prior work in imitation learning and inverse reinforcement learning has estimated the quality of demonstrations to extrapolate to better performance \citep{brown_extrapolating_2019, chen_learning_2020, zhang_confidence-aware_2022, cao_learning_2021}. \citet{beliaev_imitation_2022} estimates the expertise of demonstrators for improved imitation learning, and prior work in supervised learning estimates the expertise of human labelers to better aggregate labels from different sources \citep{ whitehill_whose_2009, raykar_learning_2010, welinder_multidimensional_2010}. However, to the best our knowledge, no prior work has investigated teacher selection in the context of reinforcement learning from human feedback, as we do here.

\section{The Expertise Problem}

\subsection{Problem Formulation}
We define expertise as a teacher's ability to reliably asses a given trajectory according to their underlying preferences. Expertise can vary across both teachers and regions of the state space. We model the teacher as a Boltzmann-rational decision-maker whose rationality varies as a function of queries. Formally, for a given teacher we define a $\beta$-function $\beta: \Sigma \times \Sigma \rightarrow \mathbb{R}$. The probability that a teacher with $\beta$-function $\beta$ selects $\sigma_1$ over $\sigma_2$ is: 

\begin{equation}
P[\sigma_1 \succ \sigma_2; \beta] = \frac{\exp(\beta(\sigma_1, \sigma_2) \cdot r(\sigma_1))}{\exp(\beta(\sigma_1, \sigma_2) \cdot r(\sigma_1)) + \exp(\beta(\sigma_1, \sigma_2) \cdot r(\sigma_2))}
\end{equation}

Teachers are defined by their $\beta$-functions $\beta_1, ..., \beta_m$, indicating their areas of specialization or levels of expertise. Critically, the RLHF algorithm doesn't have direct access to these $\beta$-functions, and must learn which teachers give the most reliable feedback on which queries. This is the \textit{expertise problem}: selecting teachers $\beta_i$ and queries $(\sigma_1, \sigma_2)$ to maximize performance against the ground-truth reward function.

\subsection{Defining Beta}
We model $\beta$-functions as gaussian kernels in concatenated mappings of trajectory space. We define a mapping function $g: (\mathcal{O} \times \mathcal{A})^k \rightarrow \mathbb{R}^g$, and for each teacher set a centroid with center $c \in \mathbb{R}^{2g}$, width $b \in \mathbb{R}^{2g}$ and scale $a \in \mathbb{R}$, such that 

\begin{equation}
\beta(\sigma_1, \sigma_2) = a \cdot \exp(|| [b ([g(\sigma_1), g(\sigma_2)] - c) ||^2)
\end{equation}

Centroids are placed evenly across the range of $g$, the dimension(s) of the state space expertise varies over. We set $a$ to 1. $b$ is set such that an agent trained from a teacher with constant $\beta (\cdot,\cdot) = \min{ \left\{ \overline{\beta} | \forall \sigma_j \in \Sigma \, \exists \beta_i \, ; \beta_i(\sigma_j, \sigma_j) = \overline{\beta} \right\}}$ reliably converges to near optimal performance. Note that under this definition there are regions of query space that have low $\beta$ values for all teachers. We call queries in these regions \textit{inter-domain queries}, as they lie outside any teacher's domain of expertise. 

\section{Experiments}
We evaluate variants of PEBBLE~\cite{lee_pebble_2021} on the expertise problem. We compare vanilla PEBBLE, where teachers are selected randomly, to PEBBLE augmented with max-$\beta$ teacher selection, and compare disagreement, similarity, and hybrid query sampling. We train each on feedback from four synthetic teachers. We use two Deepmind control suite \citep{tassa_deepmind_2018} environments, cartpole balance and walker walk, chosen for their simplicity and use as benchmarks in prior work \citep{lee_b-pref_2021}. Results are the final episode rewards averaged over 5 runs. See \ref{learning_curves} for learning curves. 

\subsection{Teacher Selection}
For each query, the RLHF algorithm must select a teacher to apply the query to (or make no selection, in which case a teacher is chosen randomly). Using the above $\beta$ function, we compare no selection (uniform) and max-$\beta$ selection. Results in Table~\ref{tab:results} show that max-$\beta$ selection yields significant performance increases over the compared to selecting teachers uniformly.

\subsection{Query Selection}

\begin{table}
    \centering
    \begin{tabular}{ccc}
    \toprule
         & Cartpole balance & Walker walk \\
         \cmidrule(lr){2-2} \cmidrule(lr){3-3}
         Unif + Dis & $396 \pm 338$ & $259 \pm 186$ \\
         Max-$\beta$ + Dis & $756 \pm 334$ & $890 \pm 95$ \\
         Max-$\beta$ + Sim & $580 \pm 262$ & $\mathbf{950} \pm 21$ \\
         Max-$\beta$ + Hybrid & $\mathbf{794} \pm 196$ & $939 \pm 71$ \\ 
    \bottomrule
    \\
    \end{tabular}
    \caption{Mean and standard deviation of final episode rewards received by PEBBLE trained with feedback from four synthetic teachers using a variety of teacher selection and query sampling methods. Across both environments, max-$\beta$ teacher selection significantly improves performance over uniform teacher selection. Similarity query sampling improves performance in walker walk, and hybrid similarity-disagreement sampling improves performance in cartpole balance. Vanilla similarity query sampling hurts performance in cartpole balance, possibly to do a bias towards uninformative queries.}
    \label{tab:results}
\end{table}

When prompting a teacher for feedback, the RLHF algorithm must select which trajectories to have the teacher compare -- the problem of query selection. Inter-domain queries are queries that have low $\beta$ values for all teachers. To reduce the probability of inter-domain queries,  we first test \textit{similarity sampling}, where queries are sampled to minimize the euclidean distance between mappings of trajectories in the query:

\begin{equation}
    \min_{(\sigma_1, \sigma_2) \in \mathcal{D}} ||g(\sigma_1) - g(\sigma_2)||
\end{equation}

To combat a bias toward uninformative queries, we propose combining similarity sampling with \textit{disagreement sampling}, which preferentially samples queries that the networks in the reward model ensemble predict different rewards for. The resulting \textit{hybrid sampling} method computes and normalizes both disagreement and similarity metrics across a batch of queries, then samples queries to maximize the difference in the normalized disagreement and similarity scores.

Results of varying sampling methods combined with max-$\beta$ teacher selection are shown in Table~\ref{tab:results}. On cartpole balance, we find similarity sampling degrades performance while hybrid sampling improves performance. On walker walk performance of all three methods is comparable, with both similarity and hybrid sampling narrowly improving over disagreement sampling.

\section{Discussion}
We model expertise as a query dependent function $\beta$ in which the error rate of teachers depends on the queries they evaluate. We find that vanilla PEBBLE is significantly outperformed by expertise-aware max-$\beta$ teacher selection on two continuous control environments from an RLHF benchmark suite. The performance gap between vanilla PEBBLE and max-$\beta$ teacher selection demonstrates the potential reward that PEBBLE and other such RLHF methods sacrifice by treating the feedback from all teachers equally, when teachers actually have varying levels of expertise. Of course, always selecting the teacher with the highest level of expertise on the current query will not be feasible with real humans, since this information is implicit in the human decision-making process and not directly accessible to the algorithm. Future work should use the formalism and implementation we have provided here to investigate methods that estimate teacher expertise and incorporate this information into teacher selection.


\textbf{Limitations and Future Work} First, while generating synthetic feedback allows us to evaluate RLHF algorithms at scale, it also requires us to manually define $\beta$-functions representing domains of expertise. Future work could replace these synthetic teacher models with real specialized humans, perhaps by using human experts on tasks that require special knowledge to evaluate, pre-training different teachers to evaluate different tasks, or artificially limiting the task information that different teachers have access to. Second, thus far we have experimented on low dimensional environments which require relatively small models. Future work could extend this analysis to more complex tasks, such as language generation. Finally, future work should investigate solutions to the expertise problem proposed here. Preliminary investigation suggests this is challenging, but promising directions include inferring teacher expertise from variation in responses to similar queries, and training neural network models of teacher expertise alongside reward models.
\bibliography{references, refs}

\pagebreak

\appendix

\section{Appendix}

\newcommand{\lccap}{Learning curves from PEBBLE trained from synthetic feedback given by four teachers with different $\beta$-functions (domains of expertise). Solid lines and shaded regions are the averages and standard deviations taken over five runs}
\subsection{Learning Curves}

\label{learning_curves}
\begin{figure}[h]
    \centering
    \includegraphics[width=15cm]{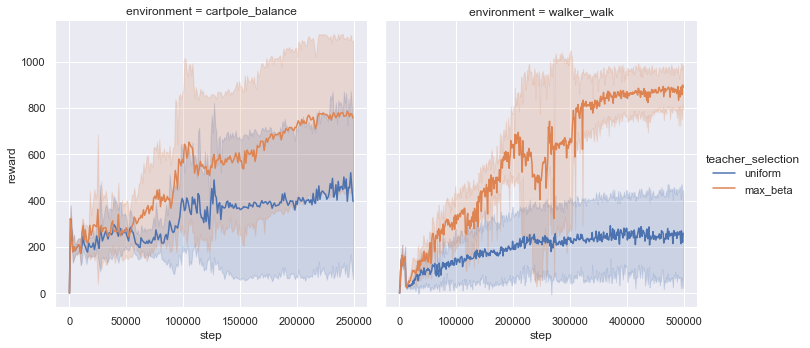}
    \caption{\textbf{Teacher Selection}: \lccap. Learning from the best available teacher (highest $\beta$) produces a large, stable performance gain over uniform teacher selection.}
     \label{fig:teach_select}
\end{figure}

\begin{figure}[h]
    \centering
    \includegraphics[width=15cm]{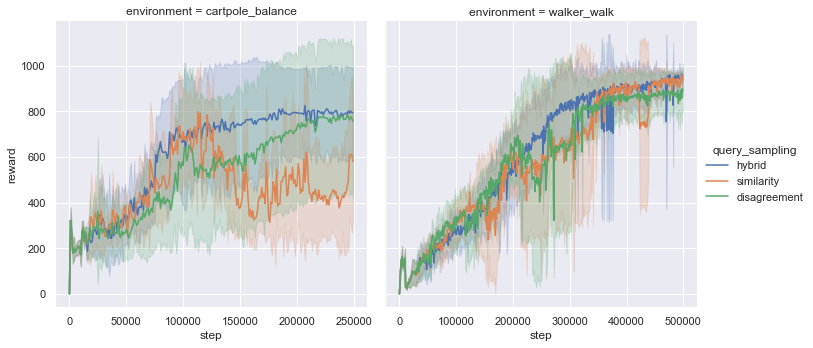}
    \caption{\textbf{Query Sampling}: \lccap. Max-$\beta$ teacher selection is used for each run. On cartpole balance, similarity sampling produces an early boost in performance which later degrades, possibly due to a bias toward uninformative queries. Hybrid sampling maintains the early boost, but stabilizes, outperforming disagreement sampling. On walker walk there are no large differences in performance, though hybrid sampling produces a particularly stable learning curve.}
    \medskip
    \small
    
     \label{fig:query_sample}
\end{figure}

\clearpage
\section{Existental Risk Analysis}
\subsection{Long-Term Impact on Advanced AI Systems}
In this section, we analyze how this work shapes the process that will lead to advanced AI systems.
\begin{enumerate}
\item \textbf{Overview.} How is this work intended to reduce existential risks from advanced AI systems? \\
\textbf{Answer:} Improved performance on the expertise problem may generically improve our capability to train ML models using human feedback (e.g.\ by increasing sampling efficiency or expanding the domain of tasks agents can learn). Given the demonstrated difficulty of specifying correct objectives~\cite{krakovna_specification_2020, clark_amodei_2016, leike_scalable_2018}, and the potentially catastrophic results of objective misspecification~\citep{christiano_what_2019}, techniques such as RLHF that allow ML systems to learn objectives instead of requiring designers to specify them may be crucial. While such techniques may also contribute to capabilities gains (as argued in \citep{hendrycks_x-risk_2022}), they may also help to keep AI capabilities on difficult-to-measure but safety-critical metrics competitive with general capabilities gains. For example, RLHF is the building block for the scalable alignment protocol ``recursive reward modeling'' \citep{leike_scalable_2018} which could ultimately be used to automate the (difficult to measure) task of alignment research \citep{leike_minimal_2022}. 
\item \textbf{Direct Effects.} If this work directly reduces existential risks, what are the main hazards, vulnerabilities, or failure modes that it directly affects? \\
\textbf{Answer:} Progress on the expertise problem reduces risk from objective misspecification and from oversight failures in dangerous domains that require expert supervision. We predict that inter- and intra-teacher variation in expertise will increase with task complexity, and therefore that this expertise problem will become increasingly relevant as we apply RLHF to nuanced, real-world tasks with the capacity for catastrophic risk. Our goal is to anticipate these problems, and create tools to facilitate research in advance. 
\item \textbf{Diffuse Effects.} If this work reduces existential risks indirectly or diffusely, what are the main contributing factors that it affects? \\
\textbf{Answer:} Formalizing the expertise problem and releasing a concrete implementation makes iterative improvement easier. If reward modeling is used in scalable alignment protocols, as proposed by~\citet{leike_scalable_2018}, improvements in reward modeling help to keep capabilities in important domains (e.g.\ alignment research) competitive with capabilities in other domains (e.g.\ maximizing profits). 
\item \textbf{What’s at Stake?} What is a future scenario in which this research direction could prevent the sudden, large-scale loss of life? If not applicable, what is a future scenario in which this research
direction be highly beneficial? \\
\textbf{Answer:} By improving reward modeling, this work may allow humans to communicate nuanced, accurate objectives to AI systems, avoiding catastrophic failures from objective misspecification. Moreover, if reward modeling is used to train AIs in specialized domains like nuclear engineering, identifying and querying teachers with relevant expertise would help prevent the AI from making catastrophic mistakes.
\item \textbf{Result Fragility.} Do the findings rest on strong theoretical assumptions; are they not demonstrated using leading-edge tasks or models; or are the findings highly sensitive to hyperparameters? \\ 
\textbf{Answer:} These results rest on the classic assumption~\cite{ziebart2010modeling, jeon_reward-rational_2020} of human Boltzmann-rationality (though they relax the assumption that the AI system knows the human's $\beta$-parameter \textit{a priori}). The problem is demonstrated using unrealistically low-dimensional continuous control tasks, chosen for their use in previous benchmarks. Finally, the experimental hyperparameters were chosen somewhat arbitrarily.
\item \textbf{Problem Difficulty.} Is it implausible that any practical system could ever markedly outperform humans at this task? \\ 
\textbf{Answer:} No. While the expertise problem is very challenging, since it requires the agent to simultaneously infer the underlying reward function and the generative process converting that reward function to observed preferences, it is not implausible that a future system may perform superhumanly well at it.
\item \textbf{Human Unreliability.} Does this approach strongly depend on handcrafted features, expert supervision, or human reliability? \\ 
\textbf{Answer:} Human supervision is naturally an integral part of our examination of learning from human feedback. However, the expertise problem is intended to expose the limitations of assuming that all humans can provide feedback reliably. In that respect, human unreliability supports our conclusions.
\item \textbf{Competitive Pressures.} Does work towards this approach strongly trade off against raw intelligence, other general capabilities, or economic utility? \\ 
\textbf{Answer:} Not to our knowledge.
\end{enumerate}
\subsection{Safety-Capabilities Balance}
In this section, we analyze how this work relates to general capabilities and how it affects the balance between safety and hazards from general capabilities.
\begin{enumerate}
\item \textbf{Overview.} How does this improve safety more than it improves general capabilities? \\
\textbf{Answer:} We hope to increase the capabilities of RLHF, in turn improving performance on hard-to-specify tasks, such as alignment research~\cite{leike_scalable_2018}. We anticipate that this will help reduce the capabilities gap between alignment-relevant tasks and easy-to-specify raw capabilities tasks. (However, note that this outcome is uncertain. See~\citet{hendrycks_x-risk_2022} for counterargument.)
\item \textbf{Red Teaming.} What is a way in which this hastens general capabilities or the onset of x-risks? \\
\textbf{Answer:} Reward modeling has improved capabilities on a number of tasks, including recent remarkable improvement in natural language generation \citep{ouyang_training_2022, askell_general_2021}. Further improvement in reward modeling may lead to further capabilities gains in these and other state-of-the-art systems, hastening existential risk from advanced and general AI systems. Moreover, reward modeling may decrease the amount of system-specific expertise required to train models, widening the pool of people who could theoretically train and apply advanced AI systems for antisocial purposes.
\item \textbf{General Tasks.} Does this work advance progress on tasks that have been previously considered the subject of usual capabilities research? \\
\textbf{Answer:} This work does not directly advance capabilities on any applied tasks. Our primary contribution is characterizing and demonstrating a challenging problem, rather than proposing a generalizable technique to solve it. Our experiments are done in small-scale continuous control environments chosen for their use in previous benchmarks~\cite{lee_b-pref_2021}.
\item \textbf{General Goals.} Does this improve or facilitate research towards general prediction, classification, state estimation, efficiency, scalability, generation, data compression, executing clear instructions,
helpfulness, informativeness, reasoning, planning, researching, optimization, (self-)supervised learning, sequential decision making, recursive self-improvement, open-ended goals, models accessing the
Internet, or similar capabilities? \\
\textbf{Answer:} This project facilitates research toward general improvements in inferring complex, compound, or otherwise difficult-to-specify objectives. This may contribute to improving agent helpfulness, researching, sequential decision making, and other tasks that are difficult-to-specify and involve interaction with humans. 


\item \textbf{Correlation With General Aptitude.} Is the analyzed capability known to be highly predicted by general cognitive ability or educational attainment? \\
\textbf{Answer:} No.
\item \textbf{Safety via Capabilities.} Does this advance safety along with, or as a consequence of, advancing other capabilities or the study of AI? \\
\textbf{Answer:} While our motivation for facilitating improvements in reward modeling is to improve safety, it may also contribute to other capability gains as discussed above.
\end{enumerate}
\subsection{Elaborations and Other Considerations}
\begin{enumerate}
\item \textbf{Other.} What clarifications or uncertainties about this work and x-risk are worth mentioning?

\textbf{Answer:} Since this work addresses anticipated problems with reward modeling, the contribution of this work to AI safety and existential risk largely depends on the contribution of reward modeling generally.  
\vspace{5pt}

In summary, progress on reward modeling reduces the capabilities gap between easy-to-define tasks and hard-to-define tasks, keeping performance on safety relevant tasks like alignment research competitive with other tasks like maximizing GDP or click through rate \citep{christiano_what_2019, leike_scalable_2018, leike_why_2022}. However, reward modeling has also lead to significant capabilities gains in applied systems such as large language models \citep{stiennon_learning_2020, askell_general_2021, ouyang_training_2022}. While these gains are often construed as ``reducing the alignment tax'', they plausibly contribute to hastening transformative AI and x-risk. 
\vspace{5pt}

While we currently expect improvements in reward modeling to be net-positive for reducing existential risk, we do have uncertainty on this point. The particulars of the expertise problem \emph{might} sway the analysis (by improving specialised oversight in safety critical domains), but this consideration is unlikely to outweigh strong views about the safety/capabilities trade off of reward modeling research.
\end{enumerate}

\end{document}